\documentclass[11pt]{article}

\usepackage{cvpr}
\usepackage{times}
\usepackage{epsfig}
\usepackage{graphicx}
\usepackage{amsmath}
\usepackage{amssymb}

\usepackage{mdwmath,algorithmic,algorithm,bm}
\usepackage[caption=false,font=footnotesize]{subfig}
\usepackage{enumerate}

\usepackage[pagebackref=true,breaklinks=true,letterpaper=true,colorlinks,bookmarks=false]{hyperref}

\cvprfinalcopy % *** Uncomment this line for the final submission

\newcommand{\Vector}[1]{\mathbf{#1}}
\newcommand{\Matrix}[1]{\mathbf{#1}}
\newcommand{\Tupel}[1]{\mathcal{#1}}

\newcommand{\R}[1]{\mathbb{R}^{#1}}

\newcommand{\sparsecoeffmatrix}{\Tupel{X}}

\DeclareMathOperator{\vvec}{vec}

\DeclareMathOperator{\tr}{tr}

\DeclareMathOperator{\SP}{{\mathrm S}}

\DeclareMathOperator{\Mani}{\mathrm{M}}
\DeclareMathOperator{\thresh}{thresh}

\newcommand{\PSNR}{\textit{PSNR}}
\newcommand{\SSIM}{\textit{SSIM}}

\def\cvprPaperID{1667} % *** Enter the CVPR Paper ID here

\ifcvprfinal\fi
\begin{document}

%%%%%%%%% TITLE
\title{Separable Dictionary Learning}

\author{Simon Hawe $\quad$ Matthias Seibert $\quad$ Martin Kleinsteuber\\
Department of Electrical Engineering and Information Technology\\
Technische Universit\"at M\"unchen
80333 M\"unchen, Germany \\
{\tt\small \{simon.hawe,m.seibert,kleinsteuber\}@tum.de} \\
%{\tt\small www.gol.ei.tum.de}
}

\maketitle
%\thispagestyle{empty}

%%%%%%%%% ABSTRACT
\begin{abstract}
Many techniques in computer vision, machine learning, and statistics rely on the fact that a signal of interest admits a sparse representation over some dictionary. Dictionaries are either available analytically, or can be learned from a suitable training set. While analytic dictionaries permit to capture the global structure of a signal and allow a fast implementation, learned dictionaries often perform better in applications as they are more adapted to the considered class of signals. 
In imagery, unfortunately, the numerical burden for (i) learning a dictionary and for (ii) employing the dictionary for reconstruction tasks only allows to deal with relatively small image patches that only capture local image information. 
%
%However, high dimensional signals and the numerical costs for applying the
%learned dictionary in reconstruction tasks pose enormous computational challenges. \\
%
%However, state-of-the-art dictionary learning methods can only deal with rather low dimensional signals, or signal patches due to the computational complexity of the learning algorithms. 
%
%Consequently, learned dictionaries are substantially different to analytic dictionaries and only cover local signal information.\\
%
%
%In this paper, we combine the advantages of fast implementation and of capturing the global structure of a signal. 
%through 
%learning the dictionary for a given class of signals.
%with the ability to learn the dictionary. 

The approach presented in this paper aims at overcoming these drawbacks by allowing a separable structure on the dictionary throughout the learning process. On the one hand, this permits larger patch-sizes for the learning phase, on the other hand, the dictionary is applied efficiently in reconstruction tasks. The learning procedure is based on optimizing over a product of spheres which updates the dictionary as a whole, thus enforces basic dictionary properties such as mutual coherence explicitly during the learning procedure. In the special case where no separable structure is enforced, our method competes with state-of-the-art dictionary learning methods like K-SVD.

%
%This is achieved by enforcing a separable structure on the dictionary throughout the learning process. Depending on the dimension of the signal, we propose two algorithms based on geometric optimization on the product of spheres. For signals of moderate dimension, we suggest a geometric conjugate gradient method, while for learning large scale dictionaries we use an adaption of stochastic gradient descent to the geometric setting. 
\end{abstract}

%%%%%%%%% BODY TEXT
\section{Introduction}
%Many successful signal reconstruction and data analysis algorithms exploit the fact that a signal $\mathbf{s} \in \R{n}$ has a sparse representation over some dictionary $\Matrix{D} \in \R{n \times d}$.
Exploiting the fact that a signal $\mathbf{s} \in \R{n}$ has a sparse representation over some dictionary $\Matrix{D} \in \R{n \times d}$ is the backbone of many successful signal reconstruction and data analysis algorithms. Having a sparse representation means that $\mathbf{s}$ is the linear combination of only a few columns of $\Matrix{D}$, referred to as \emph{atoms}. Formally,
this reads as
\begin{equation}\label{eq:dict_approx}
\mathbf{s} = \Matrix{D} \mathbf{x},
\end{equation}
where the transform coefficient vector $\mathbf{x} \in \R{d}$ is \emph{sparse}, \ie most of its entries are zero or small in magnitude. For the performance of algorithms exploiting this model, it is crucial to find a dictionary that allows the signal of interest to be represented most accurately with a coefficient vector $\mathbf{x}$ that is as sparse as possible. Basically, dictionaries can be assigned to two classes: \emph{analytic dictionaries} and \emph{learned dictionaries}. 
Analytic dictionaries are built on mathematical models of a general type of signal they should represent. They can be used universally and allow a fast implementation. Popular examples include Wavelets \cite{wav:mallat:1989}, Bandlets \cite{band:pennec:2005}, and Curvlets \cite{curv:starck:2002} among several others.
%They offer the advantages of low computational complexity and of being universally applicable to a wide set of signals. However, this universality comes at the cost of not giving the optimally sparse representation for more specific classes of signals.\\ It is now well known that signals belonging to a specific class can be represented with fewer coefficients over a dictionary that has been learned using a representative training set, than over analytic dictionaries.
It is well known that learned dictionaries yield a sparser representation than analytic ones. Given a set of representative training signals, dictionary learning algorithms aim at finding the dictionary over which the training set admits a maximally sparse representation. Formally, let $\Matrix{S} = [\mathbf{s}_1,\ldots,\mathbf{s}_m] \in \R{n \times m}$ be the matrix containing the $m$ training samples arranged as its columns, and let $\Matrix{X} = [\mathbf{x}_1,\ldots,\mathbf{x}_m] \in \R{d \times m}$ contain the corresponding $m$ sparse transform coefficient vectors, then learning a dictionary can be stated as the minimization problem
\begin{equation}\label{eq:basic_dict}
\operatorname*{minimize}_{\Matrix{X},\Matrix{D}} g(\Matrix{X}) \operatorname{~subject~to~} \|\Matrix{D} \Matrix{X} - \Matrix{S}\|_F^2 \leq \epsilon, \ \Matrix{D} \in \mathfrak{C}.
\end{equation}
Therein, $g: \R{d \times m} \to \R{}$ is a function that promotes sparsity, 
$\epsilon$ reflects the noise power, 
and $\mathfrak{C}$ is some predefined admissible set of solutions. Common dictionary learning approaches employing optimization problems related to \eqref{eq:basic_dict} include probabilistic ones like \cite{dl:engan:1999,dl:delgado:2003,dl:zhou:2009}, and clustering based ones such as K-SVD \cite{dl:aharon:2006}, see \cite{dl:tosic:2011} for a more comprehensive overview. The dictionaries produced by these techniques are unstructured matrices that allow highly sparse representations of the signals of interest. However, the dimension of the signals which are sparsely represented and, consequently, the possible dictionaries' dimensions are inherently restricted by limited memory and limited computational resources. Furthermore, when used within signal reconstruction algorithms where many matrix vector multiplications have to be performed, those dictionaries are computationally expensive to apply.

In this paper, we present a method for learning dictionaries that are efficiently applicable in reconstruction tasks.
The crucial idea is to allow the dictionary to have a separable structure, where separable means that the dictionary $\Matrix{D}$ is given by the Kronecker product of two smaller dictionaries $\Matrix{A} \in \R{h \times a}$ and $\Matrix{B} \in \R{w \times b}$, \ie
\begin{equation}\label{eq:separab}
\Matrix{D}=\Matrix{B} \otimes \Matrix{A}.
\end{equation}
The relation between a signal $\Vector{s} \in \R{hw}$ and its sparse representation $\Vector{x} \in \R{a b}$ as given in  \eqref{eq:dict_approx} is accordingly 
%
%\begin{eqnarray}
$
\Vector{s} =(\Matrix{B} \otimes \Matrix{A}) \Vector{x} =  \vvec(\Matrix{A} \vvec^{-1}(\Vector{x}) \Matrix{B}^\top),
$
%\vvec^{-1}(\Vector{s}) = \vvec^{-1}\left((\Matrix{B} \otimes \Matrix{A}) \Vector{x}\right) = \Matrix{A} \vvec^{-1}(\Vector{x}) \Matrix{B}^\top.
%\end{eqnarray}
%
where the vector space isomorphism $\vvec \colon \R{a \times b} \to  \R{a b}$ is defined as the operation that stacks the columns on top of each other.
Employing this separable structure instead of a full, unstructured dictionary clearly reduces
the computational costs of both the learning algorithm and the reconstruction tasks.
More precisely, for a separation with $h,w \sim \sqrt{n}$, the computational burden reduces from $O(n)$ to $O(\sqrt{n})$. 
We will refer to this new learning approach as \emph{SeDiL} (Separable Dictionary Learning).

It is apparent that this approach applies in principle to any class of signals. However, we will focus on signals that have an inherently two dimensional structure such as images. However, it is worth mentioning that SeDiL can straightforwardly be extended to signals with higher dimensional structure, such as volumetric $3D$-signals, by employing multiple Kronecker products. To fix the notation for the rest of this work, if $\Matrix{A}$ and $\Matrix{B}$ are as above, the two dimensional signal $\Matrix{S} \in \R{h \times w}$ has the sparse representation $\Matrix{X} \in \R{a \times b}$, \ie
%
%\begin{equation}
$\Matrix{S} = \Matrix{A} \Matrix{X} \Matrix{B}^\top$.
%\end{equation}
%
%SeDiL is based on an adaption of Problem \eqref{eq:basic_dict} to a product of unitary spheres. More precisely, we employ a Riemannian conjugate gradient method for solving the optimization problem. For the special case $\Matrix{B}=1$, this yields a new algorithm for learning standard unstructured dictionaries. We provide numerical evidence that dictionaries learned in that way can compete with those learned by state-of-the-art methods like K-SVD. For the general case, our approach is able to learn dictionaries for large patch dimensions where conventional learning techniques fail.
%%allows to learn dictionaries on patches being so large that other dictionary learning algorithms fail. 
%As an example, we learn a separable dictionary on a face database, each face image having a resolution of $64 \times 64$ pixels. To show that the separable dictionary is able to capture the global information of the underlying training set, a face inpainting experiment is given.

The proposed dictionary learning scheme SeDiL is based on an adaption of Problem \eqref{eq:basic_dict} to a product of unit spheres. Furthermore, it incorporates a regularization term that allows to control the dictionary's mutual coherence. The arising optimization problem is solved by a Riemannian conjugate gradient method combined with a nonmonotone line search. For the general separable case, the method is able to learn dictionaries for large patch dimensions where conventional learning techniques fail while if we define $\Matrix{B}=1$ SeDiL yields a new algorithm for learning standard unstructured dictionaries. A denoising experiment is given that shows the performance of both a separable and a non-separable dictionary learned by SeDiL on $(8 \times 8)$-dimensional image patches. From this experiment it can be seen that the separable dictionary outperforms its analytic counterpart, the overcomplete discrete cosine transform, and the non-separable one achieves similar performance as state-of-the-art learning methods like K-SVD. Besides that, to show that a learned separable dictionary is able to extract and to recover the global information contained in the training data, a separable dictionary is learned on a face database with each face image having a resolution of $64 \times 64$ pixels. This dictionary is then applied in a face inpainting experiment where large missing regions are recovered solely based on the information contained in the dictionary.

%The fact that it is possible to capture global information of the underlying training set is illustrated by a face inpainting experiment.

%
%
%Here, we show that it is possible to capture global information of the underlying training set. We provide a face inpainting experiment where we employ a dictionary that has been learned on face images with a resolution of $64 \times 64$ pixels.

%
%We present two methods for learning $\Matrix{A}$ and $\Matrix{B}$ based on an adaption of Problem \eqref{eq:basic_dict}. Both introduced algorithms are geometric optimization methods on a product of spheres. First, we propose a
%Patch Based Separable Dictionary Learning algorithm \emph{(PBS-DL)} that is efficiently applicable for the purpose of rapid reconstruction tasks.
%%
%Secondly, to deal with high dimensional signals, or very large training sets where memory limitations become the algorithm's bottleneck, we are suggesting a Large Scale Separable Dictionary Learning algorithm \emph{(LSS-DL)} based on a geometric stochastic gradient descent method.
%%\Dumm{The performance of both algorithms is demonstrated in several experiments.}

\section{Structured Dictionary Learning}
Instead of learning dense unstructured dictionaries, which are costly to apply in reconstruction tasks and  are unable to deal with high dimensional signals, techniques exist that aim at learning dictionaries which bypass these limitations.
%
%that exhibit certain structures to bypass the aforementioned limitations. % but still offer the advantage of having fine tuned atoms.
% This line of research is still in its early stage, and only few prior work in this direction exist. 
In the following, we shortly review some existing techniques that focus on learning efficiently applicable and high dimensional dictionaries, followed by introducing our approach.

\subsection{Related Work}
In \cite{dl:rubinstein:2010} and \cite{dl:Yaghoobi:2009}, two different algorithms have been proposed following the same idea of finding a dictionary such that the atoms themselves are sparse over some fixed analytic base dictionary. 
%Applying a sparse dictionary is computationally less demanding applying a dense dictionary. 
The algorithm proposed in \cite{dl:rubinstein:2010} enforces each atom to have a fixed number of non-zero coefficients, while the one suggested in \cite{dl:Yaghoobi:2009} imposes a less restrictive constraint by enforcing sparsity over the entire dictionary.
However, both algorithms employ optimization problems that are not capable of finding a large dictionary for high dimensional signals.
%In \cite{dl:rubinstein:2010}, a method is proposed for learning a dictionary such that each atom itself is $k$-sparse over some fixed analytic base dictionary. Applying a sparse dictionary requires less arithmetic operations than applying a dense dictionary, hence, it is computationally less demanding.  The concrete learning algorithm extends the famous K-SVD approach by introducing an additional prior that enforces the dictionary's atoms to be sparse. However, as this algorithm is based on the K-SVD method, it is still not capable of dealing with high dimensional signals.\\
%An approach called compressible dictionary learning which also follows the idea of finding a dictionary that is sparse over some base dictionary has been introduced in \cite{dl:Yaghoobi:2009}. In contrast to \cite{dl:rubinstein:2010}, global sparsity is enforced on the entire dictionary. The authors state that this model is less restrictive as compared to the method of \cite{dl:rubinstein:2010}. Again, this approach cannot deal with high dimensional signals.\\
In \cite{dl:aharon:2008} an alternative structure for dictionaries has been proposed.
The so called signature dictionary is a small image itself, where every patch at varying locations and size is a possible dictionary atom. The advantages of this structure include near-translation-invariance, reduced overfitting, and less memory and computational requirements, compared to unstructured dictionary approaches. However, the small number of parameters in this model also makes this dictionary more restrictive than other structures. This approach has been further extended in \cite{dl:benoit:2011} to learn real translational-invariant atoms.
Hierarchical frameworks for tackling high dimensional dictionary learning are presented in \cite{jenatton:10} and \cite{Zhen:11}. The latter work uses this framework in conjunction with a screening technique and random projections. 
We like to mention that our approach has the potential to be combined with 
hierarchical frameworks.

% in conjunction with random projections.

%Multiscale representations are a common structure in various analytic dictionaries such as wavelets. A dictionary learning approach following this idea is suggest in \cite{dl:mairal:2008}, where a semi-multiscale structure is employed. The resulting dictionary is a concatenation of several scale-specific subdictionaries over a dyadic grid which are learned via the K-SVD algorithm on image pyramids. The presented results show that this dictionary structure is advantageous for applications such as denoising and inpainting. However, the resulting subdictionaries are still explicit dense matrices having the drawbacks of high computational complexity.
\subsection{Proposed Approach}
We aim at learning a separable dictionary $\Matrix{D} = \Matrix{B} \otimes \Matrix{A}$ from a given set of training samples $\Tupel{S} = (\Matrix{S}_1,\dots,\Matrix{S}_m)\in \R{h \times w \times m}$ by solving a problem related to \eqref{eq:basic_dict}.
We denote the collection of the $m$ sparse representations by $\sparsecoeffmatrix=\left( \Matrix{X}_1, \dots, \Matrix{X}_m\right)$ and measure its  overall sparsity via
\begin{equation}\label{eq:lognorm1}
g(\sparsecoeffmatrix) := \sum_{j=1}^m\sum\limits_{k=1}^{a}\sum\limits_{l=1}^{b} \ln(1+\rho|x_{klj}|^2),
\end{equation}
where $x_{klj}$ is the $(k,l)$-entry of $\Matrix{X}_j \in \R{a \times b}$ and $\rho > 0$ is a weighting factor.
%
%We denote the collection of the $M$ sparse representations by $\sparsecoeffmatrix=\left( \Matrix{X}_1, \dots, \Matrix{X}_M\right)$ and measure its  overall sparsity by
%%
%\begin{equation}\label{eq:sparse}
%g(\sparsecoeffmatrix) := \tfrac{1}{q}\sum\limits_{j=1}^M p(\Matrix{X}_j)^q.
%\end{equation}
%%
%The exponent $q>1$ ensures that sparsity is distributed over all coefficient vectors. In this way, we find a dictionary which is applicable to a broad set of signals, and not overfitted to some specific subset.
%
%
We impose the following regularization on the dictionary.
%
%\compress
\begin{enumerate}[(i)]
\item The \emph{columns} of $\Matrix{D}$ have unit Euclidean norm. %i.e.\ $\|\mathbf{d}_{i}\|_2=1$ for $i=1,\ldots,d$.
\item The \emph{coherence} of $\Matrix{D}$ shall be moderate.%  i.e.\ $\mathbf{d}_i \neq \pm \mathbf{d}_j$ for $i \neq j$.
\end{enumerate}
Constraint (i) is commonly employed in various dictionary learning procedures to avoid the scale ambiguity problem, \ie the entries of $\Matrix{D}$ tend to infinity, while the entries of $\Tupel{X}$ tend to zero as this is the global minimizer of the unconstrained sparsity measure $g(\Tupel{X})$. Matrices with normalized columns admit a manifold structure, known as the product of spheres, which we denote by
\begin{equation}
\SP(n,d) := \{\Matrix{D} \in \R{n \times d} | \operatorname*{ddiag}(\Matrix{D}^\top \Matrix{D})=\Matrix{I}_d\}.
\end{equation}
Here, $\operatorname*{ddiag}(\Matrix{Z})$ forms a diagonal matrix with the diagonal entries of the square matrix $\Matrix{Z}$, and $\Matrix{I}_d$ is the $(d \times d)$-identity matrix.
Consequently, we require that $\Matrix{A}$ is an element of  $\SP(h,a)$ and that $\Matrix{B}$ is an element of $\SP(w,b)$.

The soft constraint (ii) of requiring a moderate mutual coherence of the dictionary is a well known regularization procedure in dictionary learning, and is motivated by the compressive sensing theory. Roughly speaking, the mutual coherence of $\Matrix{D}$
measures the similarity between the dictionary's atoms, or, ''\emph{a value that exposes the dictionary's vulnerability, as [...] two closely related
columns may confuse any pursuit technique.}'' \cite{elad:07}. 
The most common mutual coherence measure for a dictionary $\Matrix{D}$ with normalized columns $\mathbf{d}_i$ is
\begin{align}
\label{eq:mc_classical}
\mu(\Matrix{D}):=\max_{i < j} |\mathbf{d}_i^\top \mathbf{d}_j|.
\end{align}
For the rest of this paper we will follow this notation and denote the $i^\mathrm{th}$ column of a matrix $\Matrix{Q}$ by the corresponding lower case character $\mathbf{q}_i$.
In order to relax this worst case measure, other measures have been introduced in the literature that are more suited for practical purpose, for example averaging the largest entries of 
$\{ |\mathbf{d}_i^\top \mathbf{d}_j| ~|~ i < j \}$ as in \cite{CS:donoho:2003,elad:07,Tropp:04}, or by considering the sum of squares of all elements in $\{ |\mathbf{d}_i^\top \mathbf{d}_j| ~|~ i < j \}$, cf. \cite{duarte:09}. 
In this work, we introduce an alternative mutual coherence measure, which has been proven extremely useful in our experiments. Explicitly, we measure the mutual coherence via
\begin{align}
\label{eq:mc_new}
r(\Matrix{D}):= - \hspace{-4mm} \sum \limits_{1 \leq i < j \leq d} \ln(1-(\mathbf{d}_{i}^\top \mathbf{d}_{j})^2).
\end{align}
Since this measure is differentiable, it can be integrated into smooth optimization procedures. Furthermore, when it is used within a dictionary learning scheme, the log-barrier function avoids the algorithm from producing dictionaries that contain repeated identical atoms.

Note that minimizing $r(\Matrix{D})$ implicitly influences $\mu(\Matrix{D})$. Concretely, the relation between \eqref{eq:mc_new} and the classical mutual coherence \eqref{eq:mc_classical} is 
\begin{align}\label{eq:relation}
r(\Matrix{D}) \geq -\ln(1-(\mu(\Matrix{D}))^2) \geq \tfrac{1}{N} r(\Matrix{D}),
\end{align} 
with $N:=d(d-1)/2$ denoting the number of summands of \eqref{eq:mc_classical}.
To see the validity of the above equation, note that since the atoms $\mathbf{d}_{i}$ are normalized to one, the equation $0 \leq | \mathbf{d}_{i}^\top \mathbf{d}_{j}|^2  \leq 1$ holds due to the Cauchy-Schwarz Inequality. Thus, all summands $-\ln(1-(\mathbf{d}_{i}^\top \mathbf{d}_{j})^2)$ are non-negative. 
Moreover,
\begin{align}
\max_{i<j} (-\ln(1-(\mathbf{d}_{i}^\top \mathbf{d}_{j})^2)) = -\ln(1-(\mu(\Matrix{D}))^2),
\end{align}
and therefore 
\begin{align}
 -N \ln(1-(\mu(\Matrix{D}))^2)\geq r(\Matrix{D})\geq -\ln(1-(\mu(\Matrix{D}))^2)
\end{align}
which implies Equation \eqref{eq:relation}. 
In order to exploit this relation for the separable case we first consider the following Lemma.

\textbf{Lemma 1.} \emph{The mutual coherence of the Kronecker product of two matrices  $\Matrix{A}$ and $\Matrix{B}$ with normalized columns is equal to the maximum of the individual mutual coherences, \ie
\begin{equation}
	\mu(\Matrix{B}\otimes\Matrix{A}) = \max\{\mu(\Matrix{A}),\mu(\Matrix{B})\}.
\end{equation}}

\textsl{Proof.} First, notice that since the columns of $\Matrix{A}$ and $\Matrix{B}$ all have unit norm, the diagonal entries of both $\Matrix{A}^\top\Matrix{A}$ and $\Matrix{B}^\top \Matrix{B}$ are equal to one and that the mutual coherence $\mu(\Matrix{A})$ and $\mu(\Matrix{B})$ is given by largest off-diagonal absolute value of $\Matrix{A}^\top\Matrix{A}$ and $\Matrix{B}^\top \Matrix{B}$, respectively.
Analogously, $\mu(\Matrix{B} \otimes \Matrix{A})$ is just 
the largest off-diagonal absolute value of the matrix $(\Matrix{B} \otimes \Matrix{A})^\top (\Matrix{B}\otimes\Matrix{A}) = (\Matrix{B}^\top\Matrix{B}) \otimes (\Matrix{A}^\top\Matrix{A})$. 
Due to the definition of the Kronecker product and the unit diagonal, each entry of $\Matrix{B}^\top\Matrix{B}$ and $\Matrix{A}^\top\Matrix{A}$ reappears in the off-diagonal entries of $(\Matrix{B} \otimes \Matrix{A})^\top (\Matrix{B}\otimes\Matrix{A})$. This yields the two inequalities $\mu(\Matrix{B}) \leq \mu(\Matrix{B}\otimes\Matrix{A})$ and $\mu(\Matrix{A}) \leq \mu(\Matrix{B}\otimes\Matrix{A})$, which can be combined to
\begin{equation}
	\label{eq:cohgeq}
	\max\{\mu(\Matrix{A}),\mu(\Matrix{B})\} \leq \mu(\Matrix{B}\otimes\Matrix{A}).
\end{equation}

On the other hand, each entry of $(\Matrix{B}^\top\Matrix{B}) \otimes (\Matrix{A}^\top\Matrix{A})$ is a product of entries of $\Matrix{B}^\top\Matrix{B}$ and $\Matrix{A}^\top\Matrix{A}$. This explicitly means that we can write $\mu(\Matrix{B} \otimes \Matrix{A}) = \tilde{b}\,\tilde{a}$ with $\tilde{b}$ and $\tilde{a}$ being entries of $\Matrix{B}^\top\Matrix{B}$ and $\Matrix{A}^\top\Matrix{A}$, respectively. Since we have $0 \leq \tilde{a},\tilde{b} \leq 1$, this provides the two inequalities $\mu(\Matrix{B} \otimes \Matrix{A}) \leq \tilde{b}$ and $\mu(\Matrix{B} \otimes \Matrix{A}) \leq \tilde{a}$, and hence
\begin{equation}
	\label{eq:cohleq}
	\mu(\Matrix{B}\otimes\Matrix{A}) \leq \max\{\mu(\Matrix{A}),\mu(\Matrix{B})\}.
\end{equation}
Combining \eqref{eq:cohgeq} and \eqref{eq:cohleq} provides the desired result.\hfill$\square$

Substituting $\mu(\Matrix{B} \otimes \Matrix{A})$ into Equation \eqref{eq:relation} and then applying Lemma 1 yields
\begin{equation}\begin{split}
	\max \{r(\Matrix{B}),\, r(\Matrix{A})\} \geq -\ln(1-\mu(\Matrix{B}\otimes\Matrix{A})^2)\\
	\geq \max \{\tfrac{1}{N_B}r(\Matrix{B}),\, \tfrac{1}{N_A}r(\Matrix{A})\}
\end{split}\end{equation}
due to the monotone behavior of the logarithm.
Therefore, if $\max\{r(\Matrix{B}),\, r(\Matrix{A})\}$ is small, $\mu(\Matrix{B}\otimes\Matrix{A})$ is bounded as well. 
Now, in order to keep the mutual coherence of $\Matrix{B}\otimes\Matrix{A}$ moderate, we use the relation
\begin{align}
\nonumber C_1(r(\Matrix{B}) + r(\Matrix{A})) & \leq \max\{r(\Matrix{B}),\, r(\Matrix{A})\} \\ 
& \leq C_2(r(\Matrix{B}) + r(\Matrix{A})),
\end{align}
for some positive constants $C_1, C_2$
and
minimize the sum $r(\Matrix{B}) + r(\Matrix{A})$ instead of $\max\{r(\Matrix{B}),\, r(\Matrix{A})\}$ 
for computational convenience. 

Finally, putting all the collected ingredients together, to learn a separable dictionary  our goal is to minimize
\begin{align}
\label{eq:cost_func} \nonumber
f\colon &   \R{ a\times b \times m} \times \SP(h,a) \times \SP(w,b) \to \R{}, \\ \nonumber
 & (\sparsecoeffmatrix, \Matrix{A}, \Matrix{B}) \mapsto \tfrac{1}{2 m}\sum\limits_{j=1}^m \|\Matrix{A}\Matrix{X}_j\Matrix{B}^\top-\Matrix{S}_j\|_F^2 + \tfrac{\lambda}{m}  g(\sparsecoeffmatrix) \\ 
 &\qquad\qquad\qquad + \kappa  r(\Matrix{A}) + \kappa r(\Matrix{B}).
\end{align}
Therein, $\lambda \in \R{+}$ weighs between the sparsity of $\sparsecoeffmatrix$ and how accurately $\Matrix{A}\Matrix{X}_j\Matrix{B}^\top$ reproduces the training samples. Using this parameter, SeDiL can handle both perfect noise free training data as well as noisy training data. The second weighting factor $\kappa \in \R{+}$ controls the mutual coherence of the learned dictionary.

\section{Learning on Matrix Manifolds}\label{sec:optim}
Knowing that the feasible set of solutions to Problem \eqref{eq:cost_func} is restricted to a smooth manifold allows us to apply methods from the field of geometric optimization to learn the dictionary. To provide the necessary notation, we shortly recall the required concepts of optimization on matrix manifolds. For an in-depth introduction on optimization on matrix manifolds, we refer the interested reader to \cite{mani:absil:2008}.

Let $\Mani$ be a smooth Riemannian submanifold of some Euclidean space, and let $f\colon\Mani \to \R{}$ be a differentiable cost function. We consider the problem of finding
\begin{equation}\label{eq:optmani}
\operatorname*{arg~min}_{\Tupel{Y}\in \Mani} f(\Tupel{Y}).
\end{equation}
To every point $\Tupel{Y} \in \Mani$ one can assign a tangent space $T_\Tupel{Y}\Mani$, which is a real vector space containing all possible directions that tangentially pass through $\Tupel{Y}$. An element $\Xi \in T_\Tupel{Y}\Mani$ is called a tangent vector at $\Tupel{Y}$. Each tangent space is associated with an inner product inherited from the surrounding Euclidean space which we denote by $\langle\cdot,\cdot\rangle$ and the corresponding norm by $\|\cdot\|$. The Riemannian gradient of $f$ at $\Tupel{Y}$ is an element of the tangent space $T_\Tupel{Y}\Mani$ that points in the direction of steepest ascent of the cost function on the manifold. For the case where $f$ is globally defined on the entire surrounding Euclidean space,
the Riemannian gradient $\Tupel{G}(\Tupel{Y})$ is simply the orthogonal projection of the (standard) gradient $\nabla f(\Tupel{Y})$ onto the tangent space $T_\Tupel{Y}\Mani$, which reads as
\begin{equation}\label{eq:rgrad}
\Tupel{G}(\Tupel{Y})=\Pi_{T_\Tupel{Y}\Mani}(\nabla f(\Tupel{Y})).
\end{equation}
A \emph{geodesic} is a smooth curve $\Gamma_{\Mani}(\Tupel{Y},\Xi,t)$ emanating from $\Tupel{Y}$ in the direction of $\Xi \in T_\Tupel{Y}\Mani$, which locally describes the shortest path between two points on $\Mani$. Intuitively, it can be interpreted as the generalization of a straight line to a manifold. The Riemannian exponential mapping, which maps a point from the tangent space to the manifold, is defined as
\begin{equation}
\exp_\Tupel{Y}\colon T_\Tupel{Y}\Mani \to \Mani, \quad \Xi \mapsto \Gamma_{\Mani}(\Tupel{Y},\Xi,1).
\end{equation}
The geometric optimization method proposed in this work is based on iterating the following line search scheme.
Given the iterate $\Tupel{Y}^{(i)}$, a search direction $\Tupel{H}^{(i)} \in T_{\Tupel{Y}^{(i)}}\Mani$, and the step size $\alpha^{(i)} \in \R{}$ at the $i^\textrm{th}$ iteration, the new iterate lying on $\Mani$ is found via
\begin{equation}\label{eq:update}
\Tupel{Y}^{(i+1)} = \Gamma_{\Mani}(\Tupel{Y}^{(i)},\Tupel{H}^{(i)}, \alpha^{(i)}),
\end{equation}
\ie following the geodesic emanating from $\Tupel{Y}^{(i)}$ in the search direction $\Tupel{H}^{(i)}$ for the length $\alpha^{(i)}$.
In the following, we concretize the above concepts for the situation at hand and
present all ingredients that are necessary to implement the proposed geometric dictionary learning method. The given formulas regarding the geometry of $\SP(n,d)$ are derived e.g.\ in \cite{mani:absil:2008}. Here we are considering the product manifold $\Mani := \R{a \times b \times m} \times \SP(h,a) \times \SP(w,b)$, which is a Riemannian submanifold of $\R{a \times b \times m} \times \R{h \times a} \times \R{w \times b}$, and an element of $\Mani$ is denoted by $\Tupel{Y}=(\sparsecoeffmatrix,\Matrix{A},\Matrix{B})$.
The tangent space at $\Matrix{D} \in  \SP(n,d)$ is given by
\begin{equation}\label{eq:oproj_sp}
T_\Matrix{D}\SP(n,d) = \{\Xi \in \R{n \times d}|\operatorname{ddiag}(\Matrix{D}^\top\Xi) = {\bm 0}\},
\end{equation}
and the orthogonal projection of some matrix $\Matrix{Q} \in \R{n \times d}$ onto the tangent space reads as
\begin{equation}
\Pi_{T_\Matrix{D}\SP(n,d)}(\Matrix{Q}) = \Matrix{Q} - \Matrix{D}\operatorname{ddiag}(\Matrix{D}^\top\Matrix{Q}).
\end{equation}
Due to the product structure of $\Mani$, the tangent space of $\Mani$ at a point $\Tupel{Y} \in \Mani$ is simply the product of all individual tangent spaces, \ie $T_\Tupel{Y}\Mani := \R{a \times b \times m} \times T_{\Matrix{A}}\SP(h,a) \times T_{\Matrix{B}}\SP(w,b)$. Consequently, in accordance with Equation \eqref{eq:oproj_sp} the orthogonal projection of some arbitrary point $\Tupel{Q}=(\Tupel{Q}_1,\Matrix{Q}_2,\Matrix{Q}_3) \in \R{a\times b \times m} \times \R{h \times a} \times \R{w \times b}$ onto the tangent space $T_\Tupel{Y}\Mani$ is
\begin{equation}\label{eq:oproj}
\Pi_{T_\Tupel{Y}\Mani}(\Tupel{Q}) = (\Tupel{Q}_1,\Pi_{T_{\Matrix{A}}\SP(h,a)}(\Matrix{Q}_2),\Pi_{T_{\Matrix{B}}\SP(w,b)}(\Matrix{Q}_3) ).
\end{equation}
Each tangent space of $\Mani$ is endowed with the Riemannian metric inherited from the surrounding Euclidean space, which for two points $\Tupel{R}=(\Tupel{R}_1,\Matrix{R}_2,\Matrix{R}_3)$ and $\Tupel{P}=(\Tupel{P}_1,\Matrix{P}_2,\Matrix{P}_3) \in T_\Tupel{Y}\Mani$ is given by
\begin{equation}
\begin{split}
&\langle\Tupel{R},\Tupel{P} \rangle := \\
&\sum_{j=1}^m\tr\left((\Matrix{R}_{1,j})^\top\Matrix{P}_{1,j}\right)+\tr(\Matrix{R}_2^\top\Matrix{P}_2)+\tr(\Matrix{R}_3^\top\Matrix{P}_3).
\end{split}
\end{equation}
%
%\Dumm{Regarding geodesics, note that in general a geodesic is the solution of a second order ordinary differential equation, meaning that for arbitrary manifolds, its computation as well as computing the parallel transport is not feasible. Fortunately, in our case the exponential mapping allows for an efficient implementation.}

The final required ingredient is a way to compute geodesics. While in general there is no closed form solution to the problem of finding a certain geodesic, the case at hand allows for an efficient implementation.
Let $\mathbf{d} \in \mathrm{S}^{n-1}$ be a point on a sphere and $\mathbf{h} \in T_\mathbf{d}\mathrm{S}^{n-1}$ be a tangent vector at $\mathbf{d}$, then the geodesic in the direction of $\mathbf{h}$ is a great circle
\begin{equation}\label{eq:mapping}
	\gamma(\mathbf{d},\mathbf{h}, t)=
	\begin{cases}
		\mathbf{d}, & \textit{if } \|\mathbf{h}\|_2= 0\\
\mathbf{d}\cos(t \|\mathbf{h}\|_2)+\mathbf{h} \tfrac{\sin(t \|\mathbf{h}\|_2)}{\|\mathbf{h}\|_2}, & \textit{otherwise.}
	\end{cases}
\end{equation}
Using this, the geodesic through $\Matrix{D} \in \SP(n,d)$ in the direction of $\Matrix{H} \in T_\Matrix{D}\SP(n,d)$ is simply the combination of the great circles emerging from each column of $\Matrix{D}$ in the direction of the corresponding column of $\Matrix{H}$, \ie
\begin{equation}
\Gamma_{\SP(n,d)}(\Matrix{D},\Matrix{H}, t) = [\gamma(\mathbf{d}_{1},\mathbf{h}_{1},t),\ldots,\gamma(\mathbf{d}_{d},\mathbf{h}_{d},t)].
\end{equation}
Now, let $\Tupel{H}=(\Tupel{H}_1,\Matrix{H}_2,\Matrix{H}_3) \in T_{\Tupel{Y}}\Mani$ be a given search direction. Due to the product structure of $\Mani$ a geodesic on $\Mani$ is given by
\begin{align}
\label{eq:mapping_mtx}
\Gamma_{\Mani}&(\Tupel{Y},\Tupel{H}, t) = \\
&(\sparsecoeffmatrix + t\Tupel{H}_1,\Gamma_{\SP(h,a)}(\Matrix{A},\Matrix{H}_2, t),\Gamma_{\SP(w,b)}(\Matrix{B},\Matrix{H}_3, t)).\nonumber
\end{align}
The shorthand notation $\Tupel{G}^{(i)}:=\Tupel{G}(\Tupel{Y}^{(i)})$ will be used throughout the rest of this paper to denote the Riemannian gradient at the $i^\textrm{th}$ iterate.

%\section{Moderate Scale Conjugate Gradient Method}
\section{Separable Dictionary Learning (SeDiL)}
To solve optimization problem \eqref{eq:cost_func}, we employ a geometric conjugate gradient (CG) method, as it offers superlinear rate of convergence, while still being applicable to  large scale optimization problems with acceptable computational complexity. Therein, the initial search direction is equal to the negative Riemannian gradient, \ie $\Tupel{H}^{(0)}=-\Tupel{G}^{(0)}$. In the subsequent iterations, $\Tupel{H}^{(i+1)}$ is a linear combination of the gradient $\Tupel{G}^{(i+1)}$ and the previous search direction $\Tupel{H}^{(i)}$. Since addition of vectors from different tangent spaces is not defined, we need to map $\Tupel{H}^{(i)}$ from $T_{\Tupel{Y}^{(i)}}\Mani$ to $T_{\Tupel{Y}^{(i+1)}}\Mani$. This is done by the so-called parallel transport $\mathcal{T}_{\Mani}(\Xi,\Tupel{Y}^{(i)},\Tupel{H}^{(i)},\alpha^{(i)})$, which transports a tangent vector $\Xi \in T_{\Tupel{Y}^{(i)}}\Mani$ along the geodesic $\Gamma_{\Mani}(\Tupel{Y}^{(i)},\Tupel{H}^{(i)},t)$ to the tangent space $T_{\Tupel{Y}^{(i+1)}}\Mani$. %For the considered manifold $\Mani$, 
Similar to the way we derived a closed form solution for the geodesic, we consider the geometry of $\SP(n,d)$ at first. The parallel transport of a tangent vector $\bm{\xi} \in T_\mathbf{d}\mathrm{S}^{n-1}$ along the great circle $\gamma(\mathbf{d},\mathbf{h},t)$ is
\begin{align}
%\begin{split}
\label{eq:patrus}
\tau(\bm{\xi},\mathbf{d},\mathbf{h},t) =\\
\bm{\xi}-\tfrac{\bm{\xi}^\top \mathbf{h}}{\|\mathbf{h}\|_2^2}( & \mathbf{d}\|\mathbf{h}\|_2\sin(t\|\mathbf{h}\|_2)+ \mathbf{h}(1-\cos(t\|\mathbf{h}\|_2))),\nonumber
%\end{split}
\end{align}
and the parallel transport of $\bm{\Xi} \in T_\Matrix{D}\SP(n,d)$ along the geodesic $\Gamma_{\SP(n,d)}(\Matrix{D},\Matrix{H},t)$ is given by
\begin{equation}\begin{split}
\label{eq:patrus_mtx}
&\mathcal{T}_{\SP(n,d)}(\bm{\Xi},\Matrix{D},\Matrix{H},t) = \\
&\ \ [\tau(\bm{\xi}_{1},\mathbf{d}_{1},\mathbf{h}_{1},t),\ldots,\tau(\bm{\xi}_{d},\mathbf{d}_{d},\mathbf{h}_{d},t)].
\end{split}\end{equation}
Thus, a tangent vector $\Xi=(\Xi_1,\bm{\Xi}_2,\bm{\Xi}_3) \in T_{\Tupel{Y}}\Mani$ is transported in the direction of $\Tupel{H} \in T_{\Tupel{Y}}\Mani$ via
\begin{equation}\begin{split}
\label{eq:patrus_mtx2}
&\mathcal{T}_{\Mani}(\Xi,\Tupel{Y},\Tupel{H},t) = \\
&\ \ (\Xi_1,\mathcal{T}_{\SP(h,a)}(\bm{\Xi}_2,\Matrix{A},\Matrix{H}_2,t),\mathcal{T}_{\SP(w,b)}(\bm{\Xi}_3,\Matrix{B},\Matrix{H}_3,t)).
\end{split}\end{equation}
Now, using the shorthand notation $\mathcal{T}^{(i+1)}_{\Xi}:=\mathcal{T}_{\Mani}(\Xi,\Tupel{Y}^{(i)},\Tupel{H}^{(i)},\alpha^{(i)})$,
%\end{eqnarray}
the new search direction is computed by
\begin{equation}\label{eq:sdir}
\Tupel{H}^{(i+1)} = -\Tupel{G}^{(i+1)} + \beta^{(i)}\mathcal{T}^{(i+1)}_{\Tupel{H}^{(i)}}.
\end{equation}
We update $\beta^{(i)}$ following the hybrid optimization scheme which is proposed in \cite{cg:dai:2001} and has shown excellent performance in practice. The authors combine the Hestenes-Stiefel (HS) and Dai-Yuan (DY) update formulas, which are given by
\begin{equation}
\label{eq:hs}
\beta^{(i)}_{\textit{HS}}  =  \tfrac{\langle\Tupel{G}^{(i+1)},\Matrix{Z}^{(i+1)} \rangle}{\langle \mathcal{T}^{(i+1)}_{\Tupel{H}^{(i)}},\Matrix{Z}^{(i+1)} \rangle},\ 
\beta^{(i)}_{\textit{DY}}  =  \tfrac{\langle \Tupel{G}^{(i+1)},\Tupel{G}^{(i+1)}\rangle}{\langle \mathcal{T}^{(i+1)}_{\Tupel{H}^{(i)}},\Matrix{Z}^{(i+1)} \rangle},
\end{equation}
with $\Matrix{Z}^{(i+1)} := \Tupel{G}^{(i+1)}-\mathcal{T}^{(i+1)}_{\Tupel{G}^{(i)}}$, to create the hybrid update formula 
\begin{equation}
\label{eq:hyb}
	\beta^{(i)}_{hyb} = \max \{ 0, \min \{ \beta^{(i)}_{\textit{HS}}, \beta^{(i)}_{\textit{DY}} \} \}.
\end{equation}

In order to find an appropriate step size $\alpha^{(i)}$, we propose a Riemannian adaption of the nonmonotone line search algorithm proposed in \cite{ls:zhang:2004}. Like other nonmonotone line search schemes it has the potential to improve the likelihood of finding a global minimum as well as to increase the convergence speed, cf. \cite{ls:dai:2002}. In contrast to the standard Armijo rule and standard nonmonotone schemes, which generally use the function value at the previous iterate or the maximum of the previous $m$ iterates, this particular method utilizes a convex combination of all function values at previous iterations. 
The pseudo code for a version of this line search scheme that is adapted to our geometric optimization problem can be found in Algorithm \ref{al:nlsa}.
\begin{algorithm}
\caption{Nonmonotone Line Search on $\Mani$ in the $i^\textrm{th}$ Iteration} \label{al:nlsa}
\begin{algorithmic}
\STATE \hspace{-4.4mm} \textbf{Input:} $t_0^{(i)}>0,\ 0<c_1<1 ,0 < c_2 < 0.5,\ \mu > 0,$ $0 \leq \eta^{(i)} \leq 1,\ Q^{(i)},\ C^{(i)}$
\STATE \hspace{-4.4mm} \textbf{Set:} $t \leftarrow t_0^{(i)}$
\WHILE{$f(\Gamma_{\Mani}(\Tupel{Y}^{(i)},\Tupel{H}^{(i)},  t )) > C^{(i)} + c_2 t \langle \Tupel{G}^{(i)},\Tupel{H}^{(i)} \rangle$}
	\STATE $t \leftarrow c_1 t$
\ENDWHILE
\STATE \hspace{-4.4mm} \textbf{Set:} $Q^{(i+1)} \leftarrow \eta^{(i)}Q^{(i)} + 1$,\\ 
\hspace{3.5mm}$C^{(i+1)} \leftarrow \left(\eta^{(i)} Q^{(i)} C^{(i)} + f(\Gamma_{\Mani}(\Tupel{Y}^{(i)},\Tupel{H}^{(i)}, t)\right)/Q^{(i+1)}$,\\
\hspace{3.5mm}$\alpha^{(i)} \leftarrow t$
\STATE \hspace{-4.4mm} \textbf{Output:} $\alpha^{(i)}$, $Q^{(i+1)}$, $C^{(i+1)}$
\end{algorithmic}
\end{algorithm}
The line search is initialized with $C^{(0)} = f(\Tupel{Y}^{(0)})$ and $Q^{(0)}=1$. Finally, our complete method of learning a dictionary with separable structure is summarized in Algorithm \ref{al:learning1}.

%%%%%%%%%%%%%%%%%%%%%%%%%%%%
% old backtracking:%%%%%%%%%
%%%%%%%%%%%%%%%%%%%%%%%%%%%%
%\begin{wrapfigure}{L}{0.5\textwidth}
%\begin{minipage}{0.5\textwidth}
%\begin{algorithm}
%\caption{Backtracking Line Search on $\Mani$} \label{al:backtrack}
%\begin{algorithmic}
%\STATE \hspace{-4.4mm} \textbf{Input:} $t_0^{(i)} > 0,\; 0 < c_1 < 1,\; 0 < c_2 < 0.5$, $\;\Tupel{Y}^{(i)},\Tupel{G}^{(i)},\Tupel{H}^{(i)}$
%\STATE \hspace{-4.4mm} \textbf{Set:} $t \leftarrow t_0^{(i)} $
%\WHILE{$f(\Gamma_{\Mani}(\Tupel{Y}^{(i)},\Tupel{H}^{(i)}, t)) > f(\Tupel{Y}^{(i)}) + t c_2 \langle \Tupel{G}^{(i)},\Tupel{H}^{(i)}\rangle$}
%\STATE $t \leftarrow c_1 t$
%\ENDWHILE
%\STATE \hspace{-4.4mm} \textbf{Output:} $\alpha^{(i)} \leftarrow t $
%\end{algorithmic}
%\end{algorithm}
%\end{minipage}
%\end{wrapfigure}
%In our implementation we empirically chose $c_1 = 0.9$ and $c_2 = 10^{-2}$. As an initial guess for the step size at the first CG-iteration $i=0$, we choose $t^{(0)}_0 =\|\Tupel{G}^{(0)}\|^{-1}$,
%as proposed in \cite{cg:gilbert:1992}. In the subsequent iterations, the backtracking line search is initialized by the previous step size divided by the line search parameter, i.e.\ $t^{(i)}_0 = {\alpha^{(i-1)}}/{c_1}$. Our complete approach for learning a patch based separable dictionary is summarized in Algorithm \ref{al:learning1}.
%
\begin{algorithm}
\caption{Separable Dictionary Learning (SeDiL)}
\label{al:learning1}
\begin{algorithmic}
\STATE \hspace{-4.4mm} \textbf{Input:} Initial dictionaries $\Matrix{A}^{(0)} \in \SP(h,a),\Matrix{B}^{(0)}\in \SP(w,b)$, training data $\Tupel{S}\in \R{h\times w\times m}$, parameters $\rho,\lambda,\kappa,\thresh$
\STATE \hspace{-4.4mm} \textbf{Set:}  $i \leftarrow 0$, $\Tupel{Y}^{(0)} \leftarrow (\{\Matrix{A}^{(0)}\Matrix{S}_k\Matrix{B}^{(0)\top})\}_{k=1}^m,\Matrix{A}^{(0)},\Matrix{B}^{(0)})$, $\Tupel{H}^{(0)}\leftarrow-\Tupel{G}^{(0)}$
\REPEAT
\STATE $\alpha^{(i)},\ Q^{(i+1)},\ C^{(i+1)}$ according to Algorithm \ref{al:nlsa} in conjunction with Equation \eqref{eq:cost_func}\vspace{1mm}
\STATE $\Tupel{Y}^{(i+1)} \leftarrow \Gamma_{\Mani}(\Tupel{Y}^{(i)},\Tupel{H}^{(i)}, \alpha^{(i)})$, cf.\ \eqref{eq:mapping_mtx}\vspace{1mm}
\STATE $\Tupel{G}^{(i+1)} \leftarrow \Pi_{T_{\Tupel{Y}^{(i+1)}}\Mani}(\nabla f(\Tupel{Y}^{(i+1)}))$, cf.\ \eqref{eq:oproj}\vspace{1mm}
\STATE $\Tupel{H}^{(i+1)} \leftarrow$ $-\Tupel{G}^{(i+1)} + \beta^{(i)}_{hyb}\mathcal{T}^{(i+1)}_{\Tupel{H}^{(i)}}$, cf.\ \eqref{eq:sdir}, \eqref{eq:hyb}\vspace{1mm}
\STATE $i \leftarrow i+1$
\UNTIL{$\|\Tupel{G}^{(i)}\| < \thresh$ $\lor \ i = $ maximum $\#$ iterations}
\STATE \hspace{-4.4mm} \textbf{Output:} $\Tupel{Y}^\star \leftarrow \Tupel{Y}^{(i)} $
\end{algorithmic}
\end{algorithm}

%%%%%%%%%%%%%%%%%%%%%%%%%%%%
%% convergence obsolete
%%%%%%%%%%%%%%%%%%%%%%%%%%%%

%In the recent paper \cite{cg:ring:2012} the convergence of a geometric CG method has been shown for the Fletcher-Reeves update formula. The proof requires 
%the so called strong Wolfe-Powell condition, i.e.
%\begin{equation}\label{eq:strongwolfepowell}
%%f(\Gamma_{\Mani}(\Tupel{Y}^{(i)},\Tupel{H}^{(i)}, t)) > f(\Tupel{Y}^{(i)}) + t c_2 \langle \Tupel{G}^{(i)},\Tupel{H}^{(i)}\rangle\\
%| \langle \Tupel{G}(\Gamma_{\Mani}(\Tupel{Y}^{(i)},\Tupel{H}^{(i)}, t)),\mathcal{T}^{(i+1)}_{\Tupel{H}^{(i)}} \rangle | \leq c_3 | \langle \Tupel{G}^{(i)},\Tupel{H}^{(i)}\rangle |,
%\end{equation}
%with $0<c_2<c_3<1$, in addition to the Armijo condition used in Algorithm~\ref{al:backtrack}.

%\textbf{Lemma 1.} \emph{Under the conditions that the Fletcher-Reeves update formula is used and that the step size fulfills the strong Wolfe-Powell condition the PBS-DL algorithm converges to a critical point, i.e.\ $\liminf_{i\to\infty}  \| \Tupel{G}^{(i)} \| = 0$.}
%
%\emph{Proof.} The considered cost function \eqref{eq:cost_func} is differentiable. Furthermore, the parallel transport induced by the Riemannian exponential mapping that is employed in our algorithm fulfills the condition
%$\| \mathcal{T}^{(i+1)}_{\Tupel{H}^{(i)}} \| \leq \|\Tupel{H}^{(i)}\|.$
%Thus, the proof presented in \cite{cg:ring:2012} applies to our problem.\hfill $\square$

\section{Experiments}
\begin{figure}
\centering
\subfloat[][\centering Unstructured Dictionary]{\includegraphics[width=0.48\linewidth]{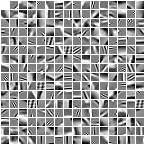}\label{fig:uatoms}}
\hfill
\subfloat[][\centering Separable Dictionary]{\includegraphics[width=0.48\linewidth]{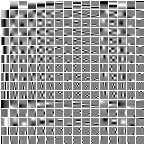}\label{fig:satoms}}
%\end{minipage}
\caption{Learned atoms of (a) unstructured dictionary $\Matrix{D_1}=1 \otimes \Matrix{A}$ and (b) separable dictionary $\Matrix{D_2}=\Matrix{B} \otimes \Matrix{A}$ for a patch size of $8 \times 8$. Each atom is shown as a $8 \times 8$ block where a black pixel corresponds to the smallest negative entry, gray is a zero entry, and white corresponds to the largest positive entry.}
\label{fig:aop}
\end{figure}

To show how dictionaries learned via SeDiL perform in real applications, we present the results achieved for denoising images corrupted by additive white Gaussian noise of different standard deviation $\sigma_\textit{noise}$ as a case study. The images and the noise levels chosen here are an excerpt of those commonly used in the literature. The peak signal-to-noise ratio ($\PSNR$) between the ground-truth image $\vvec(\Matrix{S}) \in \R{N}$ and the recovered image $\vvec(\Matrix{S}^\star) \in \R{N}$ computed by $\PSNR=10\log(255^2N/\sum_{i=1}^{N} (s_i-s_i^\star)^2)$ is used to quantify the reconstruction quality. As an additional quality measure, we use the mean Structural SIMilarity Index ($\SSIM$) computed with the same set of parameters as originally suggested in \cite{qu:wang:2004}. $\SSIM$ ranges between zero and one, with one meaning perfect image reconstruction. Compared to $\PSNR$, the $\SSIM$ better reflects the subjective visual impression of quality. 

Here, we present the denoising performance of both a universal unstructured dictionary, \ie $\Matrix{D}_1=1 \otimes\Matrix{A}$, and a universal separable dictionary $\Matrix{D}_2$, both learned from the same training data using SeDiL. By universal, we mean that the dictionary is not specifically learned for a certain image class but universally applicable to any image content. Without loss of generality we choose square image patches with $w=h=8$, which is in accordance to the patch-sizes mostly used in the literature. For the unstructured dictionary we set $a=4wh$, and for the separable one we choose $a=b=2w$, \ie $\Matrix{A}$ and $\Matrix{B}$ are of equal size and $\Matrix{D}_2 = \Matrix{B} \otimes \Matrix{A}$ is of the same dimension as its unstructured counterpart. For the training phase, we extracted $40\;000$ image patches from four images at random positions and vectorize them. Of course, these images are not considered further within the performance evaluations. The training patches were normalized to have zero mean and unit $\ell_2$-norm. We initialized $\Matrix{A}$ and $\Matrix{B}$ with random matrices with normalized columns. Global convergence to a local minimum has always been observed, regardless of the initialization. The weighting parameters were empirically set to $\rho = 100$ and $\lambda = \kappa = \tfrac{0.1}{ab}$. The resulting atoms of the unstructured dictionary $\Matrix{D}_1$ and the separable dictionary $\Matrix{D}_2 = \Matrix{B} \otimes \Matrix{A}$ are shown in Figure \ref{fig:aop}\subref{fig:uatoms} and \ref{fig:aop}\subref{fig:satoms}, respectively.

To denoise the images, we first find the sparse representation $\Matrix{X}_i^\star$ of each noisy patch $\Matrix{S}_i$ over $\Matrix{A},\Matrix{B}$ by solving
\begin{equation}\label{eq:denoise}
\Matrix{X}_i^\star = \operatorname*{arg~min}_{\Matrix{X}_i\in\R{a\times b}} \|\Matrix{X}_i\|_1 + \lambda_d \|\Matrix{A}\Matrix{X}_i\Matrix{B}^\top-\Matrix{S}_i\|_F^2.
\end{equation}
employing the Fast Iterative Shrinkage-Thresholding Algorithm (FISTA) \cite{fista:beck:2009}. The regularization parameter $\lambda_d$ depends on the noise level and we set it to $\lambda_d=\tfrac{\sigma_\textit{noise}}{100}$. After that, a clean image patch is computed from the sparse coefficients via $\Matrix{S}_i^\star=\Matrix{A}\Matrix{X}_i^\star\Matrix{B}^\top$. Last, as all overlapping image patches are taken into account, several solutions for the same pixel exist, and the final clean image is built by averaging all overlapping image patches. All achieved results are given in Table \ref{tb:psnrdenoise}.

To compare and rank the learned dictionaries among existing state-of-the-art techniques, we present the denoising performance of a universal dictionary  $\Matrix{D}_{\textit{KSVD}}$ learned using K-SVD from the same training set as used for SeDiL and of equal dimension as the unstructured dictionary $\Matrix{D}_1$. From Table \ref{tb:psnrdenoise}, it can be seen that employing $\Matrix{D}_1$ always yields slightly better denoising results compared to employing $\Matrix{D}_{\textit{KSVD}}$. Employing the separable dictionary $\Matrix{D}_2$ leads to results that are slightly worse compared to employing the unstructured counterpart. This is the tribute that has to be paid for its predefined structure. However, the separability allows a fast implementation just as 
the popular and also separable Overcomplete Discrete Cosine Transform (ODCT). Here, it can be 
observed that the separable dictionary $\Matrix{D}_2$ learned by SeDiL outperforms the ODCT for most images, while requiring exactly the same computational cost.

\begin{table*}
\centering
\caption{$\PSNR$ in dB and $\SSIM$ for denoising the five test images corrupted by five noise levels. Each cell presents the results for the respective image and noise level for five different methods: top left FISTA+K-SVD dictionary, top right FISTA+unstructured SeDiL, middle left FISTA+ODCT, middle right FISTA+separable SeDiL, bottom BM3D.}
\resizebox{\textwidth}{!}{
\begin{tabular}{||r||c|c|c|c||c|c|c|c||c|c|c|c||c|c|c|c||c|c|c|c||}
\hline  \hline
& \multicolumn{4}{c||}{\textbf{lena}} & \multicolumn{4}{c||}{\textbf{barbara}} & \multicolumn{4}{c||}{\textbf{boat}} & \multicolumn{4}{c||}{\textbf{peppers}} & \multicolumn{4}{c||}{\textbf{house}} \\ \cline{2-21}
 $\sigma_\textit{noise} \ / \ \textit{PSNR}$  & \multicolumn{2}{c|}{$\PSNR$} & \multicolumn{2}{c||}{$\SSIM$} & \multicolumn{2}{c|}{$\PSNR$} & \multicolumn{2}{c||}{$\SSIM$} & \multicolumn{2}{c|}{$\PSNR$} & \multicolumn{2}{c||}{$\SSIM$} & \multicolumn{2}{c|}{$\PSNR$} & \multicolumn{2}{c||}{$\SSIM$} & \multicolumn{2}{c|}{$\PSNR$} & \multicolumn{2}{c||}{$\SSIM$}\\
\hline  \hline
$5 \ / \ 34.15$ & 38.42 & 38.55 & 0.942 & 0.944 & 37.19 & 37.70 & 0.959 & 0.962 & 36.61 & 37.03 & 0.929 & 0.936 & 37.06 & 37.47 & 0.914 & 0.921 & 38.82 & 38.90 & 0.944 & 0.946\\\cline{2-21}
 & 38.45 & 38.51 & 0.943 & 0.946 & 37.93 & 37.65 & 0.963 & 0.965 & 37.09 & 37.04 & 0.938 & 0.938 & 37.53 & 37.39 & 0.923 & 0.922 & 39.03 & 38.90 & 0.950 & 0.948\\\cline{2-21}
 & 38.45 &  & 0.942 &  & 38.27 &  & 0.964 &  & 37.25 &  & 0.938 &  & 37.60 &  & 0.920 &  & 39.77 &  & 0.956 & \\\cline{2-21}
\hline  \hline
$10 \ / \ 28.13$ & 35.41 & 35.49 & 0.907 & 0.909 & 33.08 & 33.71 & 0.922 & 0.928 & 33.54 & 33.67 & 0.879 & 0.882 & 34.75 & 34.83 & 0.875 & 0.877 & 35.66 & 35.63 & 0.896 & 0.897\\\cline{2-21}
 & 35.29 & 35.34 & 0.907 & 0.910 & 33.99 & 33.49 & 0.931 & 0.929 & 33.45 & 33.65 & 0.879 & 0.883 & 34.65 & 34.76 & 0.876 & 0.878 & 35.37 & 35.54 & 0.896 & 0.898\\\cline{2-21}
 & 35.79 &  & 0.915 &  & 34.96 &  & 0.942 &  & 33.91 &  & 0.887 &  & 35.02 &  & 0.878 &  & 36.69 &  & 0.921 & \\\cline{2-21}
\hline  \hline
$20 \ / \ 22.11$ & 32.24 & 32.31 & 0.857 & 0.859 & 28.88 & 29.61 & 0.846 & 0.859 & 30.28 & 30.35 & 0.800 & 0.802 & 32.38 & 32.40 & 0.837 & 0.838 & 32.83 & 32.75 & 0.856 & 0.856\\\cline{2-21}
 & 32.00 & 32.11 & 0.856 & 0.858 & 29.95 & 29.28 & 0.865 & 0.854 & 29.94 & 30.25 & 0.792 & 0.800 & 31.98 & 32.23 & 0.832 & 0.838 & 32.11 & 32.45 & 0.848 & 0.854\\\cline{2-21}
 & 32.98 &  & 0.875 &  & 31.78 &  & 0.905 &  & 30.89 &  & 0.825 &  & 32.80 &  & 0.845 &  & 33.79 &  & 0.871 & \\\cline{2-21}
\hline  \hline
$30 \ / \ 18.59$ & 30.35 & 30.41 & 0.821 & 0.822 & 26.56 & 27.22 & 0.775 & 0.790 & 28.36 & 28.41 & 0.741 & 0.743 & 30.81 & 30.80 & 0.810 & 0.810 & 30.93 & 30.83 & 0.826 & 0.826\\\cline{2-21}
 & 30.02 & 30.15 & 0.817 & 0.820 & 27.61 & 26.90 & 0.800 & 0.782 & 27.96 & 28.27 & 0.729 & 0.739 & 30.28 & 30.55 & 0.803 & 0.809 & 30.07 & 30.45 & 0.815 & 0.822\\\cline{2-21}
 & 31.22 &  & 0.843 &  & 29.82 &  & 0.868 &  & 29.13 &  & 0.779 &  & 31.32 &  & 0.820 &  & 32.13 &  & 0.847 & \\\cline{2-21}
\hline  \hline
$50 \ / \ 14.15$ & 27.85 & 27.88 & 0.760 & 0.761 & 24.05 & 24.43 & 0.666 & 0.679 & 25.96 & 25.98 & 0.658 & 0.659 & 28.43 & 28.41 & 0.761 & 0.761 & 28.03 & 27.92 & 0.767 & 0.766\\\cline{2-21}
 & 27.52 & 27.64 & 0.754 & 0.758 & 24.75 & 24.24 & 0.691 & 0.671 & 25.61 & 25.83 & 0.646 & 0.654 & 27.94 & 28.18 & 0.753 & 0.759 & 27.43 & 27.60 & 0.755 & 0.760\\\cline{2-21}
 & 29.02 &  & 0.798 &  & 27.23 &  & 0.794 &  & 26.79 &  & 0.705 &  & 29.24 &  & 0.782 &  & 29.72 &  & 0.811 & \\\cline{2-21}
\hline  \hline
\end{tabular}
}
\label{tb:psnrdenoise}
\vspace{-5.5mm}
\end{table*}

\begin{figure}[htb]
\centering
%\begin{minipage}[b]{\linewidth}
  \subfloat{\includegraphics[height=0.19\linewidth]{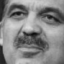}}
  \hfill
  \subfloat{\includegraphics[height=0.19\linewidth]{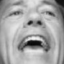}}
  \hfill
  \subfloat{\includegraphics[height=0.19\linewidth]{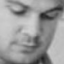}}
  \hfill
  \subfloat{\includegraphics[height=0.19\linewidth]{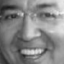}}
  \hfill
  \subfloat{\includegraphics[height=0.19\linewidth]{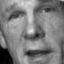}}
%\end{minipage}
\caption{Five exemplarily chosen training images.}
\vspace{-8mm}
\label{fig:faceexample1}
\end{figure}

The second advantage besides computational efficiency that comes along with the capability of learning a separable dictionary is
that SeDiL allows to learn sparse representations for image patches whose size lets other unstructured dictionary learning methods fail due to numerical reasons. In order to demonstrate the capability of SeDiL in this domain, a separable dictionary is learned from a training set consisting of $12~000$ images of dimension $(64 \times 64)$ showing frontal face views of different persons. These training images were randomly extracted from the $13~228$ faces of the "Cropped Labeled Faces in the Wild Database" \footnote{\url{http://itee.uq.edu.au/~conrad/lfwcrop/}} \cite{Huang:2007,Sanderson:2009}. The remaining $1228$ images were used for the following inpainting experiments. Note that the face positions in the pictures are arbitrary, see Figure \ref{fig:faceexample1} for five exemplary chosen training faces. The dimensions of the resulting matrices $\Matrix{A},\Matrix{B}$ were set to $(64 \times 128)$ and all other parameters required for the learning procedure were chosen as above.

%
% of $(64 \times 64)$ dimensional face images .
%
%
% and present a face inpainting example based on the dictionary. 
%  
%
%To show the capability of SeDiL to learn a separable dictionary from large image patches, and that the learned dictionary is able to represent the global information contained in the training data, we learned a dictionary from a set of $(64 \times 64)$ dimensional face images \textcolor{red}{refer to database!} and present a face inpainting example based on the dictionary. 
%
%\Dumm{Extrem homosexuell formuliert bis hier.} The training set consisted of $10~0000$ images of dimension $(64 \times 64)$ showing frontal face views of different persons. The face positions in the views are of arbitrary position, see Figure \ref{fig:faceexample} for five exemplary chosen training faces. All parameters required for learning the dictionary were chosen as mentioned above for the small scale dictionaries.
%

The ability of the separable dictionary to capture the global structure of the training samples is illustrated by an inpainting
experiment for face images of size $64 \times 64$, where large regions are missing. 
These images have of course not been included in the training set. 
We assume that the image region that has to be filled up is given.
The inpainting procedure is again conducted by applying FISTA on the inverse problem  
\begin{equation}\label{eq:inpainting}
\Matrix{X}^\star = \operatorname*{arg~min}_{\Matrix{X}\in\R{a\times b}} \|\Matrix{X}\|_1 + \lambda_d \|\mathrm{pr}(\Matrix{A}\Matrix{X}\Matrix{B}^\top)-\Vector{y}\|_2^2,
\end{equation}
where the measurements $\Vector{y} \in \R{m}$ are the available image data and $\mathrm{pr}(\cdot): \R{w \times h} \to \R{m}$ is a projection onto the corresponding region with available image data.

%where $Y$ is the available image information of the image to be inpainted and $\mathcal{A}$ the available image region.

\begin{figure}[htb]
\centering
%\begin{minipage}[b]{\linewidth}

  \subfloat{\includegraphics[height=0.19\linewidth]{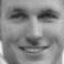}}
  \hfill
  \subfloat{\includegraphics[height=0.19\linewidth]{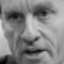}}
  \hfill
  \subfloat{\includegraphics[height=0.19\linewidth]{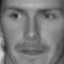}}
  \hfill
  \subfloat{\includegraphics[height=0.19\linewidth]{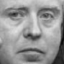}}
  \hfill
  \subfloat{\includegraphics[height=0.19\linewidth]{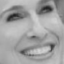}}
  \hfill 
  \vspace{-2.75mm}
  \subfloat{\includegraphics[height=0.19\linewidth]{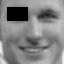}}
  \hfill 
  \subfloat{\includegraphics[height=0.19\linewidth]{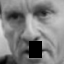}}
  \hfill
  \subfloat{\includegraphics[height=0.19\linewidth]{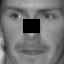}}
  \hfill
  \subfloat{\includegraphics[height=0.19\linewidth]{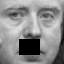}}
  \hfill
  \subfloat{\includegraphics[height=0.19\linewidth]{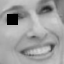}}
  \hfill 
  \vspace{-2.75mm}
  \subfloat{\includegraphics[height=0.19\linewidth]{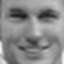}}
  \hfill
  \subfloat{\includegraphics[height=0.19\linewidth]{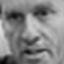}}
  \hfill
  \subfloat{\includegraphics[height=0.19\linewidth]{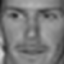}}
  \hfill
  \subfloat{\includegraphics[height=0.19\linewidth]{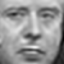}}
  \hfill
  \subfloat{\includegraphics[height=0.19\linewidth]{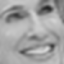}}
%\end{minipage}
\caption{Five exemplary large scale inpainting results. The first row shows the original images from which large regions are removed in the second row. The last row shows the inpainting results achieved by SeDiL.}
\vspace{-4.5mm}
\label{fig:faceexample2}
\end{figure}

An excerpt of the achieved results is given in Figure \ref{fig:faceexample2}. 
We like to mention that this experiment should not be seen as a highly sophisticated face inpainting method, but rather should supply evidence that SeDiL is able to properly extract the global information of the underlying training set.
%
%Note that this  not a highly sophisticated inpainting algorithm, but rather a simple $\ell_1$-minimization problem that shows that a separable dictionary is able to extract the information of the underlying the training set. The results will be presented soon.
%
%
%\Dumm{Wie sollen wir das korrekt aufschreiben. Optimierungsporble (35) is rein fuer denoising aber nicht fuer inpainting weil die "Messmatrix" fehlt. Ausserdem is es so nur für den separablen ansatz aufgeschrieben und nicht fuer den vektorisierten. Vielleicht faellt dir ne saubere loesung ein}
%

\section{Conclusion}
We propose a new dictionary learning algorithms called SeDiL that is able to learn both unstructured dictionaries as well as dictionaries with a separable structure. Employing a separable structure on dictionaries reduces the computational complexity from $O(n)$ to $O(\sqrt{n})$ compared to employing unstructured dictionaries, with $n$ being the considered signal dimension. Due to this, separable dictionaries can be learned using far larger signal dimensions as compared to those used for learning unstructured dictionaries, and they can be applied very efficiently in image reconstruction tasks. Another advantage of SeDiL is that it allows to control the mutual coherence of the resulting dictionary. Therefore, we introduce a new mutual coherence measure and put it in relation to the classical mutual coherence. The SeDiL algorithm we propose is a geometric conjugate gradient algorithm that exploits the underlying manifold structure. Numerical experiments for image denoising show the practicability of our approach, while the ability to learn sparse representations of large image-patches is demonstrated by a face inpainting experiment.

\subsubsection*{Acknowledgments}
This work has been supported by the Cluster of Excellence \emph{CoTeSys} - Cognition for Technical Systems, funded by the German Research Foundation (DFG).
{\small
\bibliographystyle{ieee}

}

\end{document}